\documentclass{article}
\usepackage{cite}
\usepackage{amsmath,amssymb,amsfonts}
\usepackage{algorithmic}
\usepackage{graphicx}
\usepackage{textcomp}
\usepackage{xcolor}
\usepackage{times}
\usepackage{xcolor}
\usepackage{soul}
\usepackage[utf8]{inputenc}
\usepackage[small]{caption}
\usepackage{times}
\usepackage{booktabs} % For formal tables
\usepackage{graphicx}
\usepackage{amsmath}
\usepackage{amsfonts}
\usepackage{url}
\usepackage{subfigure}
\usepackage{mathrsfs}
\usepackage[colorinlistoftodos,prependcaption,textsize=tiny]{todonotes} %todo notes
\usepackage{comment}
\usepackage[linesnumbered, ruled, vlined]{algorithm2e}
\def\BibTeX{{\rm B\kern-.05em{\sc i\kern-.025em b}\kern-.08em
    T\kern-.1667em\lower.7ex\hbox{E}\kern-.125emX}}

\begin{document}

\title{Latent Dirichlet Allocation for Internet Price War
}

% \author{
% \IEEEauthorblockN{1\textsuperscript{st} Given Name Surname}

% \title{Probabilistic Graphical Model for Internet Price War
% %\title{Winning An Internet Price War: Deep Reinforcement Learning with Latent Dirichlet Allocation for Imperfect And Incomplete Information Game

\author{Chenchen Li$^{1,2}$ \thanks{Equal contribution}, Xiang Yan$^{1,2}$ \footnotemark[1], Xiaotie Deng$^3$ \footnotemark[1], Yuan Qi$^1$ \footnotemark[1], Wei Chu$^1$ \footnotemark[1], \\
Le Song$^{1}$, Junlong Qiao$^1$, Jianshan He$^1$, Junwu Xiong$^1$\\
$^1$AI Department, Ant Financial Services Group\\
$^2$Department of Computer Science, Shanghai Jiao Tong University\\
$^3$School of Electronics Engineering and Computer Science, Peking University\\
lcc1992@sjtu.edu.cn, xyansjtu@163.com, xiaotie@pku.edu.cn\\}

\maketitle

\begin{abstract}
    Internet market makers are always facing intense competitive environment, where personalized price reductions or discounted coupons are provided for attracting more customers. Participants in such a price war scenario have to invest a lot to catch up with other competitors. However, such a huge cost of money may not always lead to an improvement of market share. This is mainly due to a lack of information about others' strategies or customers' willingness when participants develop their strategies.

    In order to obtain this hidden information through observable data, we study the relationship between companies and customers in the Internet price war. Theoretically, we provide a formalization of the problem as a stochastic game with imperfect and incomplete information. Then we develop a variant of Latent Dirichlet Allocation (LDA) to infer latent variables under the current market environment, which represents the preferences of customers and strategies of competitors. To our best knowledge, it is the first time that LDA is applied to game scenario.

    We conduct simulated experiments where our LDA model exhibits a significant improvement on finding strategies in the Internet price war by including all available market information of the market maker's competitors. And the model is applied to an open dataset for real business. Through comparisons on the likelihood of prediction for users' behavior and distribution distance between inferred opponent's strategy and the real one, our model is shown to be able to provide a better understanding for the market environment.

    Our work marks a successful learning method to infer latent information in the environment of price war by the LDA modeling, and sets an example for related competitive applications to follow.
\end{abstract}

\section{Introduction}
\setlength{\abovedisplayskip}{3pt}
\setlength{\abovedisplayshortskip}{1pt}
\setlength{\belowdisplayskip}{3pt}
\setlength{\belowdisplayshortskip}{1pt}
\setlength{\jot}{3pt}
\setlength{\textfloatsep}{3pt}
Price reduction, sometimes even below cost, is a classic tactic in market competitions, commonly referred to as a price war~\cite{kajmbt2016}.
It has become a popular strategy for the Internet platform competitions in recent years.
It is employed as a standard marketing technique to recruit participants, to boost up membership, to venture into new frontiers, and ultimately,  to eliminate competitors.

Uber, dubbed as World War Uber~\cite{ehlc2015}, has fought to conquer the world's ride-hailing market one by one in the past ten years.
Even though Uber has yet to make a profit until the third quarter 2017 ~\cite{ns2017}, its strategy of deep subsidies to attract customers has been widely adopted by the Internet companies, especially businesses built by mobile apps.
Similar tactics have been used in traditional industries such as the airlines, retails, crude oils.
But only in the context of the Internet platforms of goods and services, this strategy has become prevailing and fast evolving.
The casualties of the war are globally visible.
In the recent years, we have witnessed price wars in p2p financing, ride-hailing, bicycle sharing, online (and offline) cash back shopping~\cite{ho2017online}.
In many cases, the competition is so fierce that the number of rivals has been reduced from hundreds or thousands to a dozen or smaller, but consumed investment capitals increasing from millions to billions of dollars.
Therefore, we make it the objective of the market maker to maximize the total number of customers who would take the offer and enjoy the goods and services provided, given a fixed budget for the welfare improvement campaign.

While entrepreneurs fight to gain an advantage over opponents via financial investment, customers enjoy and benefit from the competing platforms' price wars.
In that perspectives, we view them as efforts by goods and service providers to improve the welfare of customers.
By sacrificing a portion of (future) profit, reduced prices can provide the necessities for some customers who would otherwise be not able to afford it.
%Therefore, given a fixed budget for the welfare improvement campaign (or marketing campaign), we make it the objective of the market maker to maximize the total number of customers who would take the offer and enjoy the goods and services provided.
%At the same time, we would need a customer model who takes the offers all market makers provide as input, compares with environment factors encoded in proceeding data changes, and makes the decision to take the offer or not.
%The difficulty is how to design such a framework to explore this unknown preference information and to determine what strategies other market makers take from historical data, in order to create a plausible understanding of the reality, that can be justified in practice.
However, from one company's perspective, simply setting lower prices or providing coupons worth more to customers may not always lead to more consumptions.
This is because customers have limited demand for consumptions and may have inherent preference for specific company's products.
And its opponents may also increase their investments at the same time, resulting in an equal attraction for customers.
This means on behalf of a company, providing coupons seems to have no effect on attracting customers, but in fact it will loss some customers if providing nothing to them.

Indeed, entrepreneurs' fighting in an Internet price war can be viewed as playing an imperfect and incomplete information game, to gain an advantage over opponents via financial investment.
One may distinguish between games with imperfect and perfect information through whether the opponents' potential strategies are accessible to a player.
For example, in chess or go game, each player knows all possible plays his opponent can do at any step, meaning it is a \textbf{perfection information game}.
On contrast, in card games each player's cards are often hidden from others, thus it is an \textbf{imperfect information game}, and so is the Internet price war since participant companies have no information about how their competitors provide personalized price reduction.
On the other hand, the win/loss outcome of chess, go game and card games, formally the structure of these games is known to all players after their plays, which means they are \textbf{complete information games}.
However, in an Internet price war, companies do not know how customers make their choice after receiving awards, thus not able to calculate their utilities accurately even if they know other companies' strategies.
Such a lack of customers' preference means it is also an \textbf{incomplete information game}.

If we are able to reveal these kinds of missing information, we can find the best strategy for playing such a game, and also to obtain a better understanding of the price war.
Latent Dirichlet Allocation (LDA) is a powerful tool to learn the latent variables, which have been applied in a lot of fields, such as text processing \cite{blei2003latent}, causal inference \cite{lauritzen2001causal}, image classification \cite{chong2009simultaneous} and so on.
Thus we also consider the LDA model for this scenario.
It characterizes the interactions using the observable information about consumptions in one's own company as a variable dependent on customers' preferences, which is in turn also dependent on both its strategy and its competitors' strategies of providing price reduction.
Aided by the LDA, we can infer the latent variables to approximately characterize the environment and further seek better strategies through other decision-making algorithms like Deep Reinforcement Learning (DRL).
The combined method forms a complete framework to deal with imperfect information scenario, inferring latent variables through LDA first and find better strategies based on transferred perfect information environment.

To show that the inferred information is useful in the part of decision making, we conducted experiments on simulated Internet price war game, playing against baseline methods by using our framework.
Then we apply our LDA on an open dataset from real business and evaluate the results by comparing prediction likelihood with baselines and distribution distance to the real distribution.
All these experiments justify our framework's effectiveness.

\subsection{Related Works}
Studies of price wars can be traced back to 1955 for the automobile industry \cite{bt1987}, and subsequently for airlines industry \cite{bd2017}, retail \cite{vhgepk2008}, wireless networks service \cite{mptb2010}.
Such a competitive business environment was modeled as an imperfect information game \cite{ferrero1998application}.
\cite{kajmbt2016} and \cite{rabmds2000} offer guidance for avoiding or terminating the war.
Researches also consider strategies for setting proper prices after modeling these competitions~\cite{fylblb2014,rjtj2003,wschwd2017}.
None studies micro operating strategies when a price war is inevitable.

In recent years, reinforcement learning~\cite{srba1998} is commonly believed to be useful in making strategies in game scenarios with opponents.
% Although it has not been considered for price war, reinforcement learning, or its deep learning version, DRL, has been proved to be effective on similar applications with perfect information \cite{anncsr2004,wang2016display,feng2018reinforcement,radhakrishnan2015reinforcement}.
% But when dealing with unknown information related to opponents, previous learning frameworks relied on complicated reconstruction, and their effectiveness is yet to be shown for general cases.
For example, \cite{he2016opponent} suggested an opponent modeling method adding to the action set of deep reinforcement learning.
And another famous application for imperfect information game is by \cite{heinrich2016deep}, who propose an approach named NFSP to solve the approximated Nash Equilibrium through DRL with fictitious self-play.
Their work seeks strategies under partial observed information directly but has no understanding for that unknown information.

On the other hand, exploring hidden information from observed data have been common desired in applications of data mining like recommend systems\cite{luo2016efficient}, information retrieval\cite{xuan2015infinite}, statistical natural language processing\cite{li2015generative} and so on.
Among them, probabilistic graphical models are widely used since its huge success in classifying topics from contexts\cite{blei2003latent}. Similar to our work, graphical models have been applied on inferring users' preference from user-generated data, such as \cite{giri2014user} understanding the preference of mobile device user and \cite{yu2014latent} finding buyers' preference on e-commerce search results.
But in these works, latent variables are never under the competitive environment, and as far as we know, there is no application that models one's competitor's strategies as a latent variable before this work.

\subsection{Organization of the Article}
Section 2 outlines the problem definition through a game theoretical characterization for the Internet price war.
And section 3 designs the LDA for hidden information from the environment in reality.
In Section 4 and 5, 6, we test our model systematically on simulated data and verify its suitability on a real open dataset and practical business environment.
Finally, we conclude our contributions in Section 7.

\section{Game Characterization for Internet Price War}
In this section, we formalize the Internet price war through a game theoretical characterization.
It is the first time, as far as we know, that such an important marketing phenomenon is formalized in a combined form of both macroscopical competition and microcosmic strategies.

\begin{figure}[ht]
\center

\includegraphics[width = 3.0in]{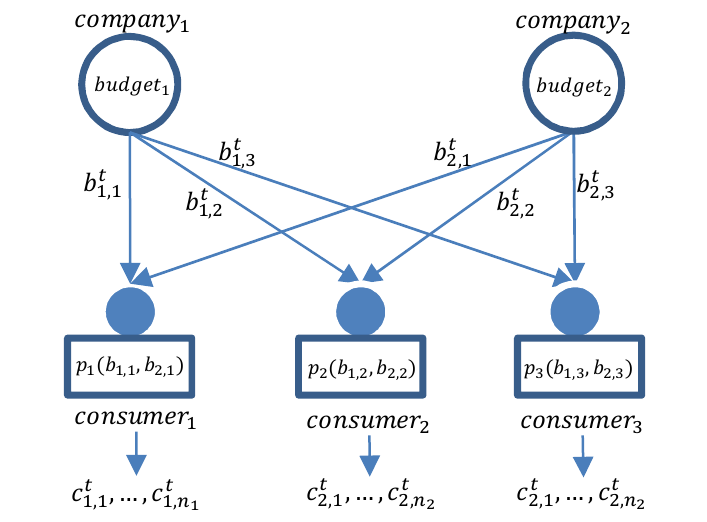}
\caption{The Internet Price War Game.}
\label{game}
\end{figure}

\subsection{Problem Definition}
As shown in Fig.\ref{game}, in an Internet price war, each company (indexed by $i=1,\dots,M$) announces personalized awards $b^{t}_{ij}$ ($b^t_{ij}\in\{0,\dots,B_i\}$, w.l.o.g $0$ means no award) for each customer in the market (indexed by $j=1,\dots,m$) during time period $(t,t+1)$ if the customer purchases its products.
Customer $j$ consumes $n^{t}_j$ times during the period, for example one week, and let $c^{t}_{j,k} = i$ if he chooses company $i$ for his $k$-th consumption.
He makes these choices according to his preference function, represented by the probability $p^{t}_j(\vec{b}^t_j,i)$ he chooses company $i$ for each consumption with respect to received awards $\vec{b}^t_j= (b^t_{1,j},b^t_{2,j},\dots,b^t_{M,j})$.

The objective for each company $i$ is to find the best strategy on providing awards $b^t_{ij}$ to maximize its market share after $R$ time periods, formally
\begin{equation}
    \label{objective}
    \max_{b^t_{ij},1\leq j\leq m, 1\leq t\leq R} \sum_{j,k,c^t_{j,k}=i} 1 / \sum_j n^t_j,
\end{equation}
under budget constraint  $\sum_j cost(b^t_{i,j}) \leq budget^t_i$, for $1\leq t\leq R$.

Corresponding to real Internet price war, each company $i$ only has its own transaction data, i.e. records of customer $j$ who received award  $b^t_{ij}$ and purchased company $i$'s products during time period $(t,t+1)$, formally $\{(t,j,b^t_{ij},c^t_{j}) | c^t_{j,k} = i\}$.
This means (1) company $i$ does not know how its opponents choose awards $b^t_{i'j}$ for $i'\neq i$, so it is playing an imperfect information game;
(2) company $i$ does not know how customers decide their consumptions $c^t_{j,k}$, so it is playing an incomplete information game.
\subsection{Basic Assumptions}
In a price war, participants are willing to provide awards for customers mainly because of two important assumptions on customers' behavior patterns:
\begin{itemize}
\item
In each short time period, say $(t,t+1)$, customers have higher probability to choose one specific company if it offers award of higher value, that is $v_i(b^t_{i,j}) > v_i(b^{'t}_{i,j})$ implies
%\begin{equation}

$$ p^t_j(b^t_{i,j},b^t_{-i,j},i) > p^t_{j}(b^{'t}_{i,j},b^t_{-i,j},i).$$

%\end{equation}
\item
After each time period, say $(t,t+1)$, the preference of customer $j$ on choosing company $i$ without any award tend to his usage rate $u_i = \sum_{k,c^{t}_{j,k} = i}1 /  \sum_{k,c^{t}_{j,k} = i'}1$ of it, that is
%\begin{equation}
 % $   (p^{t+1}_{j}(\vec{0}, i) - p^{t}_j(\vec{0},i) * (p^{t+1}_{j}(\vec{0}, i) - u_i) < 0$.
$$(p^{t+1}_{j}(\vec{0}, i) - p^{t}_j(\vec{0},i)) * (u_i - p^{t}_j(\vec{0},i)) \geq 0.$$

\end{itemize}
%\end{equation}
%\item each company $i$ only has their transaction data, which only record the customer $j$'s using company $i$.
% \item After each period, customers have higher probability to choose one specific company if he chooses it more in the last period, that is
% $$p^{t+1}_i(b^t_{i,j}, b^t_{i',j}) > p^{t}_i(b^t_{i,j}, b^t_{i',j}) ~if~\sum\limits_{k,r^{(t)}_{j,k} = i}1 >  \sum\limits_{k,r^{(t)}_{j,k} = i'}1$$
% \item
Such an evolution of customers' preference, and further evolution of related outcome function for all players in the game, make it a \textbf{stochastic game}. For the sake of analysis, we assume customers make their decisions $c^t_{j,k}$ at any time $t$ only depend on the award $b^t_{i,j}$ each company offers, but are unrelated to the total number of his consumptions $n^t_j$ in the period $(t, t+1)$, nor to other buyers' choice.
% \end{itemize}

%And for the sake of further analysis, we also need following assumptions for companies.

And for companies, since we are considering this problem from one company's perspective, all his competitors can be regarded as one opponent.
Meanwhile, as modern marketing always does, companies cluster customers into several groups, each of which contains customers of similar behavior.

Now the process of the Internet price war can be precisely described by Alg. 1.

\begin{algorithm}
        \caption{The Process of the Internet Price War}
        \label{alg:process_of_price_war}
        \KwIn{$R$, $m$, $budget_i, i \in \{ 1, 2 \}$ }
        \KwOut{The market share of each company, $s_1$, $s_2$}
        Initialize company $i, i \in \{ 1, 2 \} $ and customer $j$ with their private $v_j$ and $p^0_j$, $j \in \{1, \dots, m \}$ \\
        \For{$t \leftarrow 1$ \KwTo $R$}{
            \For{$j \leftarrow 1$ \KwTo $m$}{
                $b^t_{1, j} \leftarrow$ company 1 choose a award for customer $j$ \\
                $budget_1 \leftarrow budget_1 - cost(b^t_{1, j})$ \\
                $b^t_{2, j} \leftarrow$ company 2 choose a award for customer $j$ \\
                $budget_2 \leftarrow budget_2 - cost(b^t_{2, j})$ \\
                \For{$k \leftarrow 1$ \KwTo $n_j$}{
                    $c^t_{j,k} \leftarrow p^t_j(b^t_{1,j},b^t_{2,j})$ \\
                }
                $p^{t+1}_j \leftarrow$ update $p^{t}_j$ according to $c^{t}_{j}$
            }
        }
        $s_1 = \sum_{j, k, c^t_{j,k}=1} 1 / \sum_j n^t_j$ \\
        $s_2 = \sum_{j, k, c^t_{j,k}=2} 1 / \sum_j n^t_j$ \\
\end{algorithm}
%\subsection{Comparison with traditional price war}
%The game we introduced before also cover the situation of traditional price war. The main differences between traditional price war and Internet price war is that in traditional price war, companies offer each customer the same value of awards, while in Internet price war, they are able to offer personalized award for each customer. Personalized awards are more accurate and effective. For example, we can offer more bonus to the customers rarely buying their products while offer less bonus to the ones often buying their products.

\section{Latent Dirichlet Allocation for Price War Game}\label{lda_model}

% The strategy learning process is done through a deep reinforcement learning approach, and a PGM is designed to infer those hidden information in the game.
% The framework is shown in Fig.\ref{fig:framework}.
% Since the preference functions of the customers and the opponent's strategies would change over time, our framework is evolvable in response to the changing of the environment.
% To be specific, we separate the price war into a series of time periods.
% After each period, we adopt LDA to learn the preference distributions and competitor's strategy distribution from the transaction data of one company, then train the DRL on the updated history data to decide actions in the next period.
% The training data are the transitions between the states of the same customer in the adjacent time period.

% \subsection{Probabilistic Graphical Model for Hidden Information}

\begin{figure}[ht]
\center
\includegraphics[width = 3.0in]{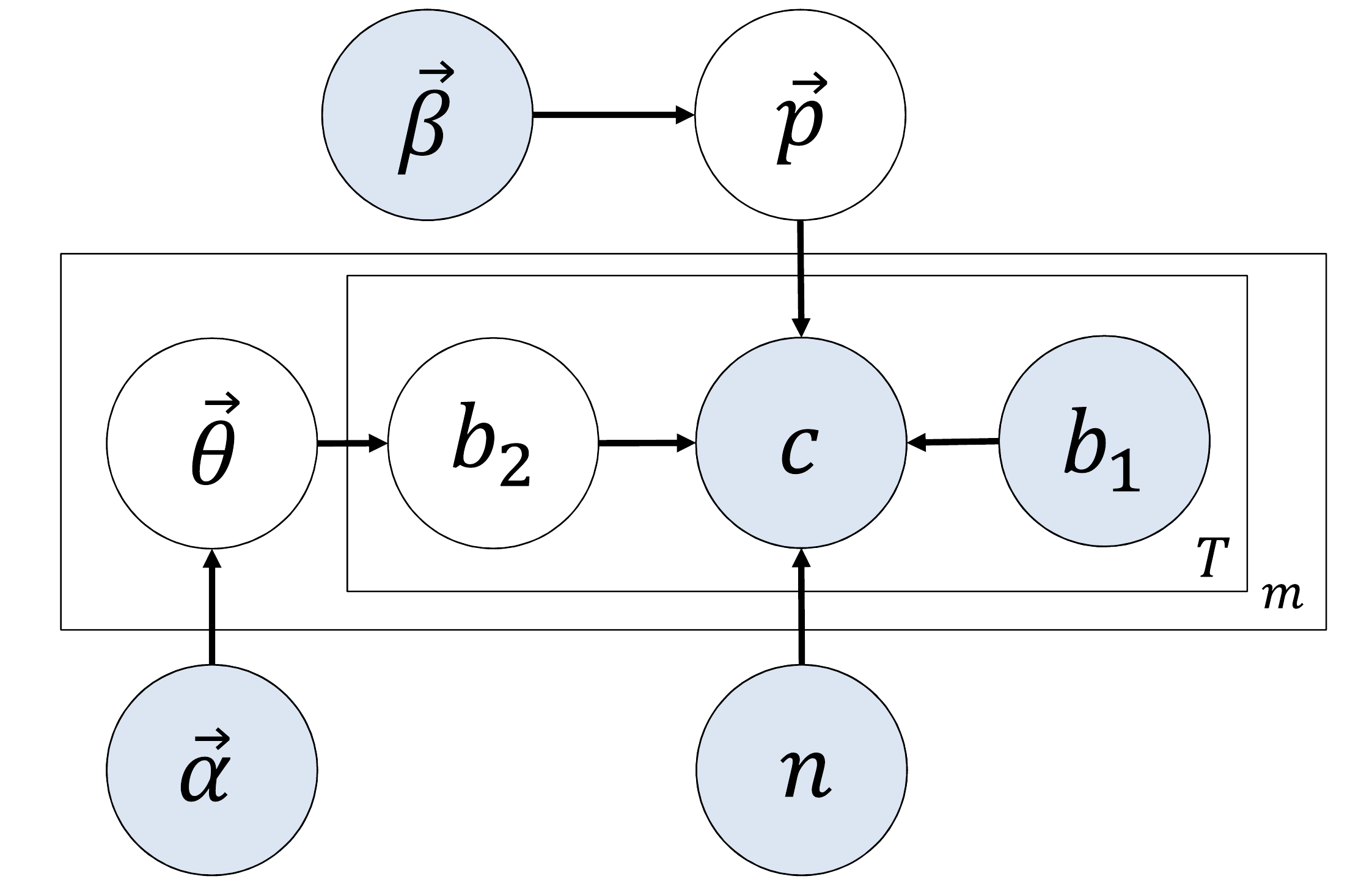}
\caption{The Internet Price War PGM.}
\label{fig:pricewar_pgm}
\end{figure}

We model the process of each customer choosing company $1$ to consume, called the Internet Price War LDA, as shown in Figure \ref{fig:pricewar_pgm}.
We omitted the superscripts about time and subscripts about customers for expressions of all variables.

\subsection{Price War LDA}
In this subsection, we first show the generative process of observed data in the game of price war, then we introduce the details.
\begin{itemize}
    \item Choose a preference distribution $\vec{p} \sim Dir(\beta)$
    \item For the each customer $j$, choose a strategy distribution $\vec{\theta} \sim Dir(\alpha)$
    \item Customer $j$ decides to conumse $n_j$ times
    \item Company 2 chooses an award $b_{2} \sim \vec{\theta}$
    \item Company 1's choice of the award $b_1$ is known
    \item For each consumption $c_{k, j}$, $k=1,2,\dots,n_j$,
        \begin{itemize}
            \item customer $j$ chooses the company $c_{k,j} \sim \vec{p}(b_1, b_2)$
        \end{itemize}
        \item Company 1 observes that customer $j$ has $c$ consumptions, where $c = \sum_{k, c_{k,j}=1} 1$

\end{itemize}

At the beginning of each time period, company $1$ decides to provide customer $j$ with award $b_1$, while his opponent company $2$ provide $b_2$.
Company $2$ decide $b_2$ according to some strategy $\vec{\theta}$, representing the probabilistic distribution of all possible awards, but the exact award $b_2$ and the distribution $\vec{\theta}$ is unknown to company $1$.
Meanwhile, customer $j$'s preference function is simplified as $\vec{p} = p(b_1, b_2)$ as the probabilistic distribution on choosing company $1$ to consume with respect to all possible awards pair $(b_1,b_2)$.
And in this period, the customer plans to consume $n$ times in total, which is subject to a distribution of $\vec{n}$.
For each consumption, he chooses one specific company according to the preference distribution $\vec{p}$ along with actually received awards $(b_1,b_2)$, thus company $1$'s observed records $c$ consist of his all consumptions on it in the period.
Since we focus on the probability that customer $j$ will choose company $1$, we consider the $\frac{c}{n}$ as observed data.
For $\frac{c}{n}$ is in [0, 1], we define a function $g(\cdot)= \llcorner \frac{c}{n} * acc \lrcorner$ to discretize their value into a new range according to required accuracy $acc$. It is noticeable that we figure out the the distribution of $\vec{n}$ by statistics in advance, rather than inferring it by LDA. When we infer the latent variables, we sample $n$ till $c \leq n$ in order to avoid that $n < c$.

Without loss of generality, we assume the hidden variables $\vec{b_2}$ and $\vec{p}$ is from two Dirichlet distribution $Dir(\vec{\alpha})$ and $Dir(\vec{\beta})$.
We define $\vec{p}$ as the multinomial distribution on the $\{0, acc-1 \}$ with size of $|B_1|*|B_2|$, where $|B_1|$ is the number of awards company $1$ provides and $|B_2|$ is the number of awards we assume the opponent offers.
And we define $\vec{b_2}$ as the multinomial distribution on the $\{0, \dots, |B_2| \}$.
%, where $|B_2|$ is the number that how many awards we assume the opponent has. % to be a power-law distribution
And on behalf of company $1$, we assume that company $2$ is using the same strategy $\vec{\theta}$ for a specific customer in recent several periods of time, say $T$.
Meanwhile, we assume company $1$ has clustered customers into groups, so that customers in each group have the same preference functions.
Thus company $1$ could use records for each group of customers in the normalized form $(j, t, b_1, g(\frac{c}{n}))$, where $j\in \{1, 2, \dots, m\}$ and $t \in \{1, 2, \dots, T\}$.
Then it is able to get approximations for the distribution of his opponent's strategy $\vec{\theta}$ for each customer $j$ and the preference function $\vec{p}$ for these group of customers by solving the Price War LDA.

\subsection{Inference}
We use the Gibbs Sampling method to solve our LDA.
% Now we show how Gibbs Sampling method calculate the distributions of $\vec{b_2}$ and $\vec{p}$.
% Here let $\vec{b} = (b_1,b_2)$ as the awards pair.
% The inference:
% In order to obtain the distribution of $\vec{p}$ and $\vec{\theta}$, we first estimate the posterior distribution over $b_2$, i.e the assignment of awards.
% The sampling distribution for a count $c_i$ given the remaining opponent's awards and preference distribution is $Pr( b_{2, j, i} = k | \vec{b}_{2,-i}, \vec{p}, \vec{c}, \vec{\alpha}, \vec{\beta}, b_1, n)$, where $\vec{b}_{2,-i}$ is vectors of assignments of the opponent's awards in the collection except for the count at position $i$ in the group $j$.
% $n$ is sampled from a fixed distribution each time.
The joint probability of the opponent's bonus $\vec{b_2}$ and count $\vec{c}$ can be factored into the following:
% $$Pr(\vec{c}, \vec{b_2}|\vec{\alpha}, \vec{\beta}, \vec{n}, \vec{b}_1) = Pr(\vec{c}|\vec{b_2} , \beta) P(\vec{b_2}|\alpha)$$.
% % For the first term, by integrating out $\theta$, we obtain
% % $$Pr(\vec{b_2}|\alpha) = \prod\limits_{i=1}^{m} p(\vec{b}_{2, m} | \vec{\alpha}) = $$
% % $$Pr(\vec{c}|\vec{b_2}, \beta) = \prod\limits_{i=1}^{m} p()$$
% %
%
% The joint probability of the awards pair and consumption distribution can be factored into the following two terms:
$$ Pr(\vec{c}, \vec{b_2}|\vec{b_1}, \vec{n}, \alpha,\beta) = Pr(\vec{c}| \vec{b_1}, \vec{b_2},\vec{n},\beta)Pr(\vec{b_2}|\alpha) $$

% For the first term, for fixed $\vec{b_1}$, by integrating out $\vec{p}$, we obtain:
% $$Pr(\vec{c}|\vec{b_1}, \vec{b_2},\vec{n},\beta) =
% % \sum\limits_nPr(n|\vec{n})\cdot(
% \frac{\Gamma(N\beta)}{\Gamma(\beta)^N})^{|B_2|}\prod\limits_{j}\frac{\prod_i\Gamma(N_{i,j}+\beta)}{\Gamma(N_{j} + N\beta)}$$
% Here $\Gamma$ is the gamma function.% and $V$ is the size of the set for all possible number of consumptions.
% $Pr(n|\vec{n})$ is the probability that the number of consumptions is $n$ given its distribution.
% $|B_2|$ is the total number of possible awards from company 2.
% $N_{i,j}$ is the number of times a customer has $i$ times consumptions appeared when $b_2 = j$.
% And $N_{j}$ is the number of times when $b_2 = j$.
%
% For the second term, by integrating out $\vec{\theta}$, we obtain:
% $$Pr(\vec{b_2}| \alpha) = (\frac{\Gamma(|B_2|\alpha)}{\Gamma(\alpha)^{|B_2|}})^{S}\prod\limits_{k}\frac{\prod_j\Gamma(N_{j,k}+\alpha)}{\Gamma(N_{k}+|B_2|\alpha)}$$
% Here $S$ is the total number of samples with company 1 providing award of $b_1$.
% $N_{j,k}$ is the number of times customer $j$ receives award $b_2 = k$.
% And $N_{k}$ is the total number of times when $b_2 = k$.

Gibbs sampling will sequentially sample each variable of interest from the distribution over that variable given the current values of other variables and the data.

According to Gibbs Sampling, and letting the subscript $-i$ denote the statistic value for an variable without the $i$-th sample, the conditional posterior for $p_i$ and $b_{2, i}$ is
\begin{equation}
 \begin{split}
     &Pr(b_{2, j, i} = k | \vec{b}_{2,-i}, \vec{p}_{-i}, \vec{c}_{i}, \alpha , \beta , b_1, n)                  \\
     & \propto \frac{ \{ N^{(h)}_{b_1, k} \}_{-i} + \beta }{ \{ N_{b_1, k} \}_{-i} + acc \beta)} *
     \frac{ \{ N^{ (k) }_{j} \}_{-i} + \alpha}{ \{ N_{j} \}_{-i} + |B_2| \alpha)}
 \end{split}
\end{equation}

Here $N^{(h)}_{b_1, k}$ is the number of times $g(\frac{c}{n}) = h$ when given $(b_{1}, k)$ and $N_{b_1, k}$ is the total number of records when given $(b_{1}, k)$. $N^{(k)}_{j}$ is the number of times customer $j$ receives $k$ from company 2 and $N_{j}$ is the total number of consumptions of customer $j$.
     % \frac{\vec{n}_{b_1,b_2,-i}+\vec{\beta}}{\sum_{i=1}^{S}(\vec{n}_{b_1,b_2,i}+\vec{\beta})}$$
    % $$Pr(c_i = g(\frac{c}{n}) | , b_{}) Pr(b_{2, i} = k | Pr(b|\vec{\alpha})$$
    % \propto $$
    % $$ Pr(c_i = g(\frac{c}{n})|\vec{p}, b_{2,i} = k, b_1, n) Pr(b_{2,i} = k | \vec{b} ) Pr( \vec{p} | \vec{\beta} )Pr( \vec{b} | \vec{\alpha} )$$

    % $$ \propto E[\Pi_{k}] E[\Phi_c]$$
    %
    % $$ Pr(b_{2,i} = k, c_{i} = g(\frac{c}{n}) | \vec{b_{2,-i}}, \vec{p_{i}}, \vec{c}_{-i}) = E[\Pi_{k}] E[\Phi_c]$$
% \end{equation}
% where
% $$E[\Pi_k] = \frac{\vec{n}_{k,-i}+\vec{\alpha}}{\sum_{i=1}^{D}(\vec{n}_{k,i}+\vec{\alpha})}$$
% $$E[\Phi_{b_1,b_2}] = \frac{\vec{n}_{b_1,b_2,-i}+\vec{\beta}}{\sum_{i=1}^{S}(\vec{n}_{b_1,b_2,i}+\vec{\beta})}$$

\subsection{Postprocessing}
It is worth noting when we get $\vec{p_{j_1}}(b_{1, j_1}, b_{2, j_1})$ and $\vec{p_{j_2}}(b_{1, j_2}, b_{2, j_2})$ via the different records of customer $j_1$ and customer $j_2$,
then they don't represent the distributions of the same pair of awards if $(b_{1, j_1}, b_{2, j_1}) = (b_{1, j_2}, b_{2, j_2})$.
% the same index of $\vec{p_{1}}$ and $\vec{p_{2}}$, learning from different records, may mean different pairs of actions after inference.
% Formally, we get $\vec{p_1}$ and $\vec{p_2}$ via the different records, then $\vec{p_1}(i, j)$ and $\vec{p_2}(i, j)$ don't represent the distributions of same pair of actions.
The reason is that we do not assign an exact award of opponents when inferring, but ids to represent then, thus the ids may indicate different actions in different times. In order to avoid the situation, we assume that the opponent has $|B_2|$ actions, where $v_j(x_1) < v_j(x_2)$ if $x_1, x_2 \in B_2$ and  $x_1 < x_2$.
According the Assumption 2 in Section 2.2, the expectation of the consumptions of customer $j$ on company 1 when $b_2 = x_1$ should larger than the one when $b_2 = x_2$ if $x_1 < x_2$. Thus we can sort the inferred preference distribution $\vec{p}$ accordingly when $b_2$ is fixed, then we can get all $\vec{p}$ in the same order.

\section{Simulations Experiments}
\label{simulate}
In this section, we introduce the experiments on the simulation framework to show that the distributions learned from our LDA is useful for coupon decision to achieve more market share.
Firstly we explain how we simulate the price war under a behavior evolution framework for customers.
Secondly, we introduce some methods able to utilize the distributions, like Deep Reinforcement Learning (DRL) and Dynamic Programming (DP).

\subsection{Preference Evolution Framework Settings}
To evaluate our method through numerical experiments, we design a preference evolution framework to simulate how customers act in a price war, motivated by Sethi and Somanathan \cite{srse}.

% \begin{itemize}
% \item\section{Simulations Experiments}
\textbf{Preference Function:} Here we focus on the situation when customer $j$ receiving awards $b^t_1$ and $b^t_2$ from company 1 and company 2 respectively.
At time $t = 0$ a customer has an initial preference distribution $p^0_{j}(b_1,b_2,1)$ on choosing company 1, dependent on the difference $d = v_j(b_1) - v_j(b_2)$ between the value of awards he receives from both companies.
The preference for choosing company 2 is naturally $1-p^0_{j}(b_1,b_2,1)$ and we do not mention it specifically in the followings.
We define $v_j(x) = cost(x) = x$ in our simulated experiments, and the notation for $p^t_{j}(b_1,b_2,1)$ can be simplified as $p^t_j(d)$.
The preference distribution takes the same form as a Sigmoid function except its mean value modified to customer $j$'s inherent preference for choosing company 1 when no award is provided.
That is letting $p^0_j(d) = Sig(d,\sigma)$ for $\forall d$ where
\begin{equation}
Sig(d, \sigma) = \left\{
\begin{aligned}
&\frac{\sigma}{0.5} \times (\frac{1}{1+e^{-d}} - 0.5) + \sigma ~~&if~d < 0\\
&\frac{1 - \sigma}{0.5} \times (\frac{1}{1+e^{-d}} - 0.5) + \sigma ~~&if~d > 0
\end{aligned}
\right.
\end{equation}
and $\sigma = p^t_j(0)$.
And whenever $\sigma$ is determined, the whole function can be determined.
We choose the preference function in this form because
(1) it increases monotonously as the value difference between awards from two companies increases, corresponding to Assumption 1 in Section 2.2;
(2) it accords with the property of diminishing marginal returns.

% \item
\textbf{Updating Process:} During the period $(t,t+1)$, customer $j$ consumes for $n_j^t$ times, each of which is independently subject to the preference distribution $p_j^t(d^t)$, where $d^t = v_j(b^t_1) - v_j(b^t_2)$ is the value difference of his actual received awards.
After that, we can calculate the usage rate $u_{i,j}^t$.
According to Assumption 2 in Section 2.2, we let the updating formula to be:
\begin{equation}
    p_{j}^{t+1}(0) = (u^t_{i,j} - p^t_j(0)) * \gamma + p^t_j(0)
\end{equation}
, where $\gamma$ is a parameter reflecting how sensitive the customer is to the awards, called updating rate.
Then the whole preference distribution can be calculated accordingly as $p^{t+1}_j(d) = Sig(d,p_{j}^{t+1}(0))$ for $\forall d$.

% \end{itemize}
\subsection{Some Methods can utilize the information}\label{sub:DRL}
In this subsection, we introduce some methods which take the advantage of the distributions learned from our model.

\subsubsection{Deep Reinforcement Learning}
Deep Reinforcement Learning is a flexible framework for Markov Decision Process.
The input of DRL only requires a fixed-length vector, which usually represents the state of the observed environment.
Thus we directly combine the preference distributions and strategy distributions with the raw features vectors.
DRL also pays attention to model the transitions between different states, which may be a good model for the evolution of customers' preferences and the transformation of the opponents' strategies.
It is also a framework of optimization, thus we do not need other extra operations.
Thus, we design a DRL framework to utilize the information of LDA, as followed:
\begin{figure*}
\center
\includegraphics[height=1.8in]{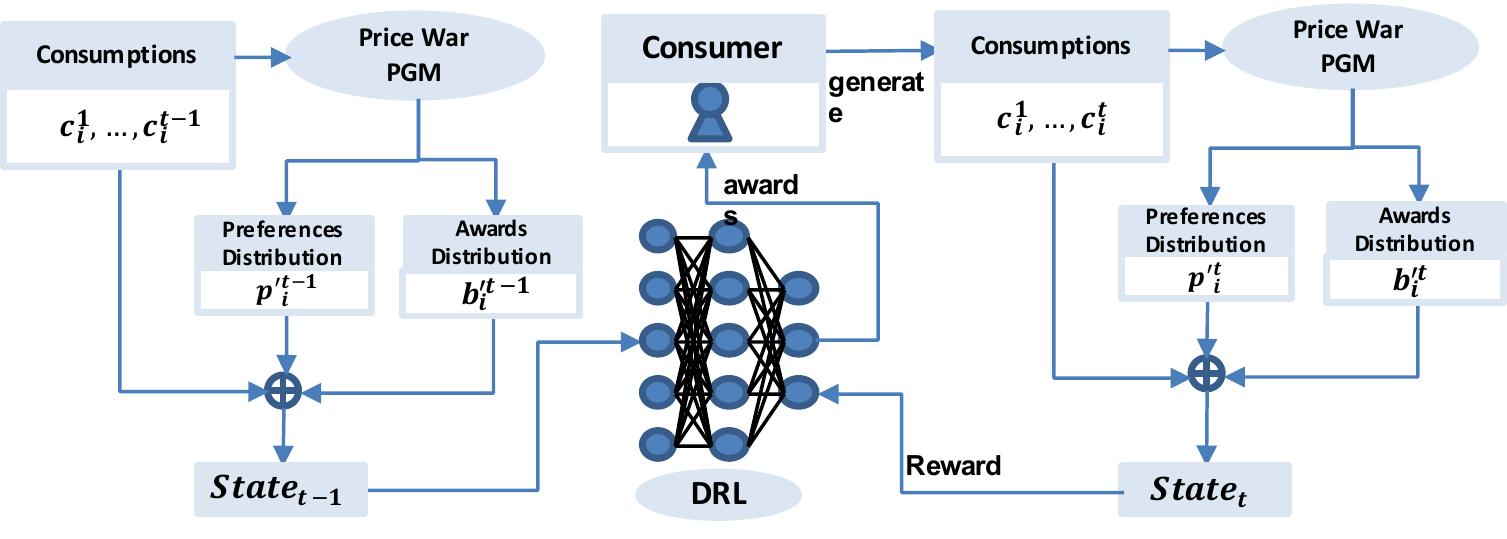}
\caption{The DRL framework for Pirce War with the information learned from the Pirce War LDA}
\label{fig:framework}

\end{figure*}

\textbf{State}:
 % \textsl{State} $s^t_i$ is a characterization for customer $i$ at the time $t$.
 $s^t_j$ contains three parts, consumptions history $h^t_j$ of customer $j$ before time $t$, preference distribution $p^{'t-1}_j$ and award distribution $b^{'t-1}_j$ of the opponents learning from $h^t_j$, which are the approximation of $p^{t-1}_j$ and $b^{t-1}_j$.
As the preference and award of opponent may change litte in a short period, i.e, ($t-1$, $t$), we can consider the $p^{t-1}_j \approx p^{t}_j$ and $b^{t-1}_j \approx b^{t}_j$. Therefore, we add the preference and opponents' award of time $t-1$ into state $s^t_j$. In this paper, we simply concat three parts, that
\begin{equation}
  s_j^t = \left (
   \begin{matrix}
     h_j^t \cr
     p_j^t \cr
     ~~~b_j^{'t-1}
   \end{matrix}
  \right )
\end{equation}
The transition is from $s^{t-1}_j$ to $s^{t}_j$ for each state.
% Specific to a price war, we want to find the good strategy against the opponents for offering bonus to each specific customers, so \textsl{State} should contain features of each customer. The
% The transitions between their basic features such as number consumptions are nature, and such choice results in a model-free RL.

\textbf{Action}:
% \textsl{Actions} are all kinds of the prize the company can provide, which can be the exact value of the award or abstract identity.
The $b^t_j$, that the award we choose for customer $j$ at time $t$ is in $\{0,.., |B_1|\}$, where $B_1$ is the set of actions predefined.
% Though companies in a price war may select numerous appearance of bonus, the most important feature of a bonus is its value in money.
In our deep reinforcement learning, the \textsl{Action} only consists of all the possible value of awards in an interval pre-announced by a company.
And for the convenience of experiments, we further discretize those value.

\textbf{Reward}:
% \textsl{Reward} of a two-tuple $(s, a)$ representing the feedback of environment after executing action $a$ at state $s$.
In a price war, when a company provide the award $b^t_j$ to a customer represented by $s^t_j$, the number of consumptions he chooses the company is a nature \textsl{Reward}.
But in real marketing, such feedback should also include a factor of cost as a negative part, since companies have limit budgets.
As a result, $\textsl{Reward} = c - \xi * cost(b^t_j)$, where $c$ represents the number of consumptions and $\xi$ is the parameter to control the weight of two parts.
The reason why a company's remainder budget are not included in \textsl{State} is because the company cannot be sure how many customers it will capture after providing the award.
On contrast, the average cost of attracting a customer matters more than the total money spends in the end.

\textbf{Framework}: Fig. \ref{fig:framework} show the overall framework. We adopt the Deep-Q-Network \cite{mvkksd} as the version of DRL.
The inputs of DQN are itemized above.
The optimization process can be defined as

\begin{equation}
    Q^t(s^t_j, a) = (1 - \alpha)Q^t(s^t_j, a) + \alpha (Reward + \lambda \max_{a} Q(s^{t+1}_j, a))
\end{equation}
where $\alpha$ is the learning rate and $\lambda$ is the discount factor.

\subsubsection{Dynamic Programming}
Since we learn the preference distributions and strategy distributions, we can do optimizations directly according to these kinds of information.
% Dynamic Programming (DP) may be a good way.
In precise, we define $\operatorname{f}(i, k)$ as the maximum market shares we can get when we finish offering awards to the first $i$ customers costing $k$ budgets.
Then we take advantage of Dynamic Programming (DP) to learn the optimal result of $\operatorname{f}(m, budget)$ in every single round.
Finally, we choose the coupon corresponding to the optimal solution for each customer as our policy.

Formally, the transition equation for solving $\operatorname{f}(i, k)$ is
\begin{equation}\label{dp_eqn}
\operatorname{f}(i, k) = \max\limits_{1 \leq j \leq |B|} (\operatorname{f}(i-1, k-j) + \psi(i, j))
\end{equation}
% where $\operatorname{f}(i, k)$ means the when we finish offering coupons to $i$ customers and we have costed $k$ budgets, the maximum market shares we can get.
$\psi(i, j)$ is expected benefit from customer $i$ if we offer award $j$ to him, which is calculated by
\begin{equation}
    \psi(i, l) = \sum_{l=0}^{|B|} b_i(j) p_i(j, l)
\end{equation}
where $b_i(j)$ is the probability that the opponent choose bonus $j$ for customer $i$, and $p_i(j, h)$ is the probability that the customer $i$ choose our company if it received $j$ from us and $l$ from the opponent.
And we choose the award $j^*$ that maximizes Eq. (\ref{dp_eqn}) for the $i$-th customer.

\subsection{Other Baseline Methods}
To evaluate our model, we conduct a series of simulation experiments.
In the experiments, company $1$ uses the DRL or DP as introduced before, to play against company $2$ using the baseline method as following:
\begin{itemize}
    % \item \textbf{Fixed Strategy} A company can always provide awards of the same value for customers, or nothing if it runs out of its budget. In our simulation experiments, there are 4 kinds of fixed strategies since there are 3 kinds of awards. Note that any possible strategy which does not vary along with time is actually a combination of these 3 fixed strategies. If our method can beat all fixed strategy, it can beat all time-independent strategies.
    \item \textbf{Random Strategy} is referred to a company randomly choosing one of the possible awards for each customer with equal probability.
    \item \textbf{Deep-Q-Network(DQN)} \cite{mvkksd} is a version of DRL. Note that the settings of this DQN are exactly the same as the ones mentioned in the subsection \ref{sub:DRL} except its state does not include features about the customer's preference and opponent's strategy.
    % \item \textbf{Dynamic Programming} After inferring the customer's preference and opponent's strategy, a company can use dynamic programming to find the best awarding strategy under budget constraint for each round. This baseline is compared with the optimal strategy for simulation but it is not efficient to deploy in real business.
%It can be regarded as a kind of greedy method for each time series, without considering the evolution of customers or the opponent, compared to reinforcement learning based method.
\end{itemize}

% We test our methods on the different value of $h$, which means how many past history data we use. For example, if the round is $t$ and $h=5$, then the data from the period $(t-5, t-1)$ are our training data.

% \subsection{Parameters Settings}
% \textbf{Simulated Environment}: In the simulated environment, there are 10 kinds of customers at all, each of which has 1000 persons. The initial $\sigma$ = 0.5, updating rate $\gamma$ = 0.5. $n^t_j \in [1, 100]$ for $\forall j, t$ There are two companies in the markets at all. Each company has 4 kinds of awards, $B_1 = B_2 = \{ 0, 1, 2, 3, 4 \}$, with the same amount of budgets, $budget_1 = budget_2 = 20000$.
%
% \textbf{Learning Methods}: We adopt Deep Q-Network \cite{mvkksd} (DQN) as our DRL method. The network has 3 layers, the sizes of which are $N_{input}$, 512, 5, where $N_{input}$ is the size of input features. The reward function is $reward = c^t_j - 0.5 * cost(b^t_j)$. The learning rate is 0.01 and memory size is 200000. The reward decay rate is 0.9.
%
% \textbf{}Here since the approximation solution to LDA is two set of variables, representing customers' preference and opponent's strategy, we do experiments of adding these two features to DQN's states separately and together, and they are referred as "DQN + Preference", "DQN + Strategy" and "DQN + LDA" respectively.

\subsection{Other Settings}

Other experimental settings such as the parameters of the preference evolution framework, the parameters of the deep reinforcement learning model, and different variants are explained below:
\begin{itemize}
  \item
\textbf{Simulated Environment}: In the simulated environment, there are 10 kinds of customers at all, each of which has 1000 persons. The initial $\sigma$ = 0.5, updating rate $\gamma$ = 0.5. $n^t_j \in [1, 100]$ for $\forall j, t$ There are two companies in the markets at all. Each company has 5 kinds of awards, $B_1 = B_2 = \{ 0, 1, 2, 3, 4 \}$, with the same amount of budgets, $budget_1 = budget_2 = 20000$.
  \item
\textbf{Learning Methods}: We adopt Deep Q-Network \cite{mvkksd} (DQN) as our DRL method. The network has 3 layers, the sizes of which are $N_{input}$, 512, 5, where $N_{input}$ is the size of input features. The reward function is $reward = c^t_j - 0.5 * cost(b^t_j)$. The learning rate is 0.01 and memory size is 200000. The reward decay rate is 0.9.
  \item
\textbf{Variants}: Here since the approximation solution to LDA are two sets of variables, representing customers' preference and opponent's strategy, we do experiments of adding these two features to DQN's states separately and together, and they are referred as "DQN + P", "DQN + S" and "DQN + LDA" respectively. And the DP introduced before requires both features, it is simply referred as 'DP'.
\end{itemize}

\subsection{Results}
We list the final market shares of company 1 after 1000 rounds in Table \ref{result}. It uses the variants of our methods (DQN, DQN+P, DQN+S, DQN+LDA and DP), playing against company 2 using Random Strategy or DQN. The market share is the average value taken from 10 repeated experiments.
% And for the case when the opponent is using Fixed Strategy, the three rate is corresponding to those three kinds of strategy.

\begin{table}[ht]
% \caption{market share of Row Player against }
\begin{center}
    % \scriptsize
\begin{tabular}{|c|c|c|c|c|c|c|}
\hline
 & DQN    & DQN + P & DQN + S & DQN + LDA & DP  \\ \hline
Random                       & 58,59\% & 68.05\% & 67.26\% &69.57\% & 76.12\% \\ \hline
DQN                          & 50.22\% & 55.84\% & 56.72\% &65.16\% & 54.63\% \\ \hline

% & Random  & DQN     \\ \hline
% DQN                          & 58.59\% & 51\%    \\ \hline
% DQN + Preference             & 68.05\% & 55.84\% \\ \hline
% DQN + Strategy               & 67.26\% & 56.72\% \\ \hline
% DQN + LDA                    & 69.57\% & 65.16\% \\ \hline
% DP                           & 76.12\% & 54.63\% \\ \hline
\end{tabular}
\caption{Comparison of market share. The number in the $i$-th row is the market share of company 1 when company 1 uses the $i$-th method in the first row playing against company 2 using the $j$-th method in the first column.}
\label{result}
\end{center}

\end{table}

Generally speaking, our methods get market shares over $50\%$ when competing with Random Strategy and DQN, which do not include specific information about customers' preference and opponent's strategy.
This means that the inferred latent variables from the Price War LDA, either separately or joint together, are helpful to characterize the environment of an Internet price war.

Meanwhile, DP shows the best result when playing against Random Strategy, while DQN + LDA performs best against DQN.
This coincides with common sense as Random Strategy is not evolving along with time, which means DP can learn the optimal solution with respect to known information.
When the opponent is using a complicated method like DQN, DQN + LDA is the most effective method because it models both the transition of the evolving environment and inferred information.
.

%\textbf{DQN + LDA vs Fixed Strategy}
%
%(detailed analysis...)
%
%\textbf{DQN vs Random}
%
%Figure \ref{dqn_lda_dqn_1} (a) shows the market share of company 1 after $t$ time periods, when using different strategies competing against company 2 using Random Strategy.
%In this group of experiments, the states of these DQN at each time $t$ contains customers' consumptions records in the past $h=1$ time period.
%Since $h$ should be a hyper parameter, we list results of experiments when choosing other $h$ in Table \ref{result}.
%As we can see, after adding features of customers' preference, opponent's strategy, and both of them, DQN beats simple method of randomly choosing at a market sharing of around 67.70\%, 68.04\% and 69.85\% respectively in 1000 time periods.
%On the contrast, the simple DQN only get 60.19\% market share.
%Basically, it shows that DQN do beat simple randomly choosing strategy in an Internet price war, and our method is able to significantly improve DQN's performance.
%The information about customers' preference and opponent's strategy, either separately or joint together, are helpful to characterize the environment DQN are facing in an Internet price war.
%
%\textbf{DQN + LDA vs DQN}
\begin{figure*}
\center
\subfigure[Market share curves when playing against Random Strategy.]{
\includegraphics[width=0.9\columnwidth]{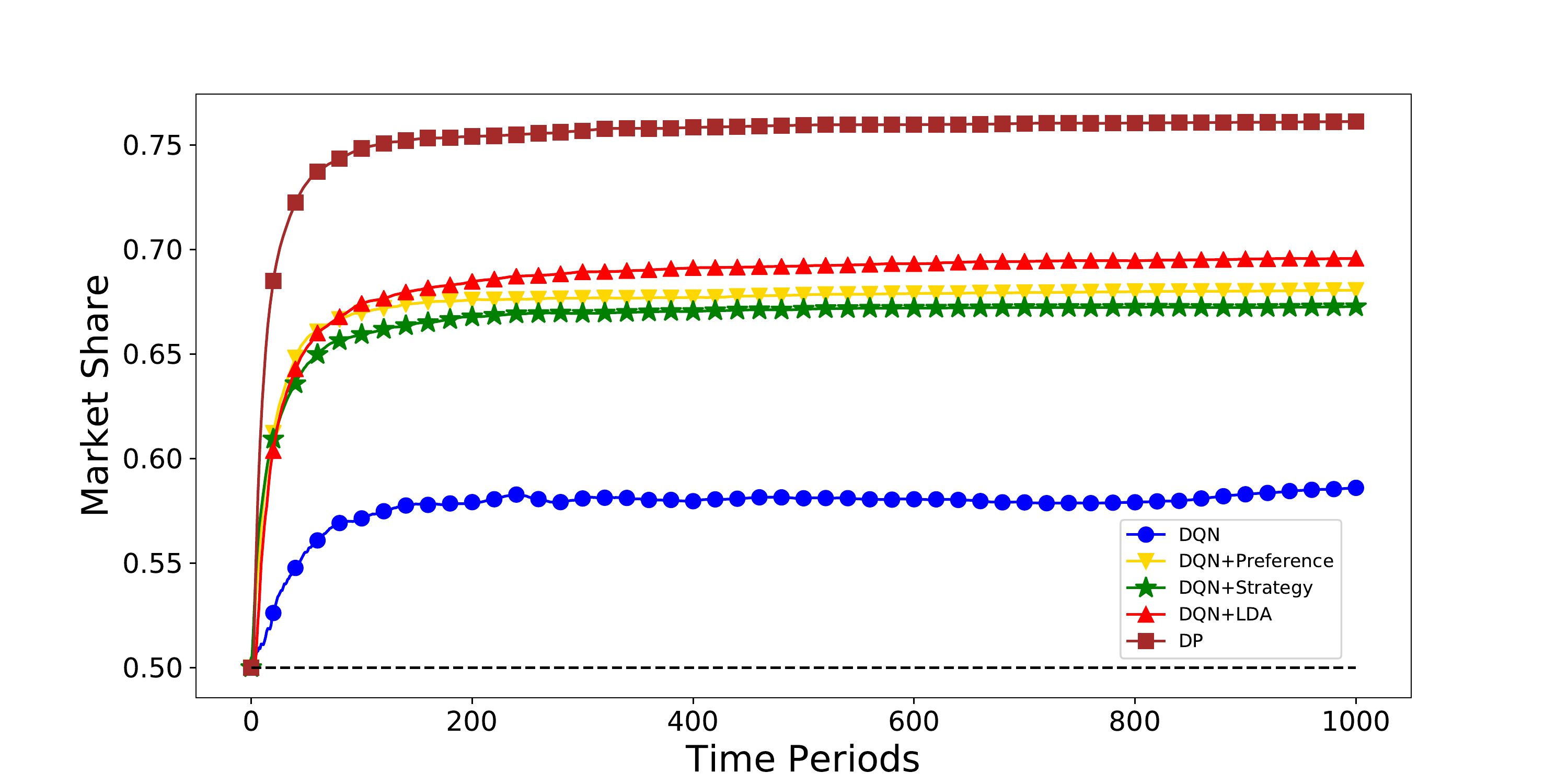}
}
\subfigure[Market share curves when playing against DQN.] { \label{fig:a}
\includegraphics[width=0.9\columnwidth]{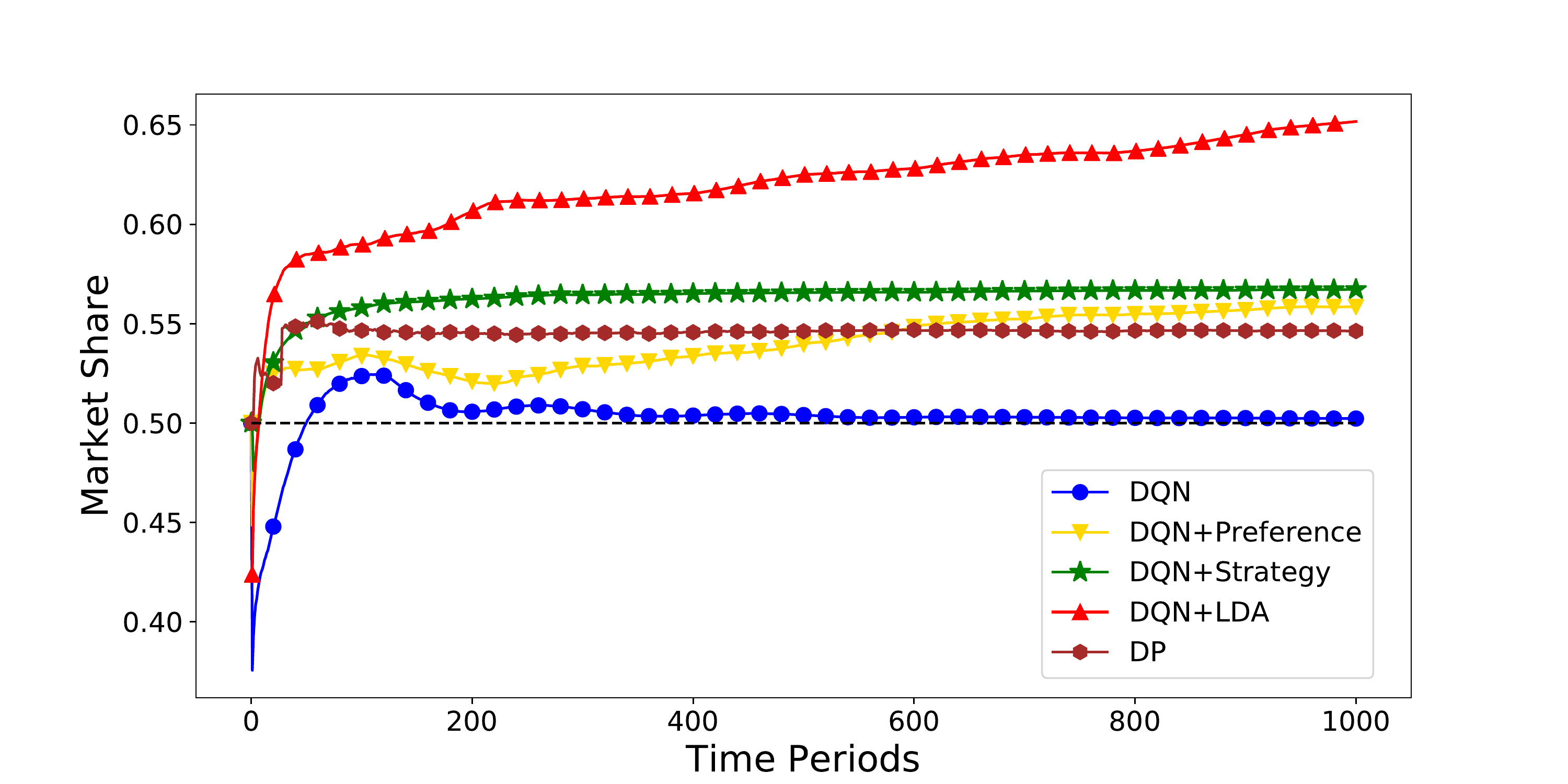}
}
% \subfigure[DQN + LDA vs DP] { \label{fig:b}
% \includegraphics[width=0.5\columnwidth]{fig/dqn_x_1_sigvs_dp_cum.pdf}
% }
\caption{Market share curves in simulation experiments.}
\label{dqn_lda_dqn_1}

\end{figure*}
% \begin{comment}
% \begin{figure}
% \center
% \includegraphics[width = 3.5in]{fig/dqn_x_3_sigvs_dqn.pdf}
% \caption{DQN + LDA with history=3 vs DQN}
% \label{dqn_lda_dqn_3}
% \end{figure}
% \begin{figure}[ht]
% \center
% \includegraphics[width = 3.5in]{fig/dqn_x_5_sigvs_dqn.pdf}
% \caption{DQN + LDA with history=5 vs DQN}
% \label{dqn_lda_dqn_5}
% \end{figure}
% \end{comment}

And Fig.\ref{dqn_lda_dqn_1} shows the average market share of company 1 after $t$ time periods, when using different strategies competing against company 2 using baseline methods.
%Similarly, these are the results when choosing $h=1$ and the other choices of $h$ are shown in Table \ref{result}.
% From the processes of these competitions, we can find that our method converges faster when the opponent is using Random Strategy.
We can find that the convergence procedures in the Fig. \ref{dqn_lda_dqn_1} (a) is faster and more stable than the ones in the Fig. \ref{dqn_lda_dqn_1} (b).
The reason is that Random Strategy can be considered as the static environment, while DQN is evolving along with the rounds.
This is in line with the intuition.

\section{Real-World Dataset Analysis}\label{o2o_data}
In the simulated experiments, the latent information inferred by our model has shown great help in finding strategies to earn more market share.
And in this section, we apply our model on a real-world dataset and conduct a series of experiments, to prove that the model can indeed infer our desired latent information.
%In the previous experiments, we evaluate our model on the simulated data to prove that we can earn more market share with the help of information learned from the latent variables model.
%However, the simulated data is generated from our theoretical model. We have to prove its effectiveness via the real-world data.
We first introduce our open dataset, followed by our preprocessing methods.
Then, we provide quantified evaluation results compared with the baselines and finally analyze the results of our method from different aspects.

\subsection{Coupon Usage Data for O2O}
\textsl{Coupon Usage Data for O2O}, referred to \textsl{O2O Dataset} in following description, is an open dataset from the Tianchi contest \cite{coupondata2018}.
O2O represents "online to offline", while a typical example of "O2O marketing" is that merchants in a shopping mall send coupons to potential customers through emails or short messages in their own APPs.
Merchants want to attract customers to their offline shops and decide these personalized discount rate of coupons based on a large amount of users' behavior and location information recorded by various APPs.

In our experiment, we make use of the offline training data from \textsl{O2O Dataset}, where the coupon promotion is conducted by 7737 retail stores from Jan. 1st 2016 to June 30th 2016.
There are 1,048,575 records in total, among which there are 255,550 users receiving 9280 kinds of coupons.
In precise, each record consists of identifications for a user, a merchant and a coupon (or 'null' if no coupon was provided), distance between the merchant and the user, the discount rate of the coupon, the date when the coupon was sent and the date when the coupon was used (or 'null' if it was never used).
After basic data cleaning and statistics, we know that on average each user receives $1.986$ coupons, while on average each kind of coupons is sent to $37.637$ customers.
Generally speaking, $7.12\%$ coupons are used.

% In this challenge, we consider the coupon promotion conducted by retail stores, of which the core is customer targeting via personal taste. People often like coupons issued by their favourite stores and feel disturbed while receiving non-related ads. Motivated by this observation, the task is to predict the probability of coupon usage in 15 days, based on their online and offline shopping data in the past. Especially, the training set is collected from Jan. 1st 2016 to June 30th 2016 while the test set from July. 1st 2016 to July 15th 2016. For privacy and commercial consideration, data are anonymized and sampled biasedly.

% \begin{table}
% \caption{statistical results of Coupon Usage Dataset}
% \begin{center}
% % \scriptsize
% \begin{tabular}{|c|c|c|c|c|}
% \hline
% Number of Users & Number of Merchant & Number of Coupons & Number of records \\ \hline
% \end{tabular}
% \end{table}

\subsection{Preprocess}\label{prepro}
Unlike the simulated data, it requires a preprocessing at first for our model to apply on the real-world data.
According to our model, it requires data for users with similar preference in each group, called preference group.
And in each group, the opponent may adopt different strategies to different people.
Since in the practice, we know neither the preferences of the users nor the strategies of the opponents, we need to cluster the data twice.
%\begin{itemize}
%    \item Firstly, we cluster the users with the similar preferences into one group, called preference group, and assume that they have the same preference distribution.
%    \item Secondly, we cluster the users in each preference group with the similar opponent's strategies distribution, called strategy group, and assume that the opponent chooses a strategy from the same distribution for all users in each specific strategy group.
%\end{itemize}
%With the above assumptions, we introduce the main steps of our preprocessing below.
Now we introduce the main steps of our preprocessing in detail.
% Therefore, at first, we need to cluster the overall users into different groups with the same preference distribution.
% Then, we still need to cluster the person in the same group into small groups again to represent that the strategy distribution in the small group is the same.
\subsubsection{Agents}
Firstly, we consider each merchant in \textsl{O2O Dataset} as an agent in such a competitive environment, providing coupons to attract customers.
In practice, each merchant is only accessible to those records related itself.
And since our LDA can help infer latent variables on behalf of one agent, we choose the merchant with id 3381 as the company 1 in our model, since it has the largest number of records in the dataset.
Then all other merchants are considered together as its opponent, in other word, the company 2 in our model.
%In our PGM, there are three agents, company 1, company 2 and customers.
%We choose the merchant with id 3381 in the dataset as the company 1 since it has the largest number of related records, and then all other merchants can be considered as one player, in another word, company 2.
%The users in the dataset are just the customers. However, some users would be considered as one user according to our assumptions, which will be introduced in details bellows.
%Specifically, in our experiment, we consider the merchant ??? as the player 1, because it has large enough data,
%While others don't meet the requirement.
%It is noticeable that the player 1 only can see the dataset related to it in our experiments.
% In this dataset, we consider one merchant as one player, $P$, and the other merchants as the other player, $O$.
% The users are just the users in all the dataset.
% When we fix one merchant as the player $P$, $P$ only can see the dataset related to it.

%detail statistic,
To be precise, there are 74823 records related to company 1, among them, 8 kinds of coupons are provided to 64152 users.
Those coupons, according to their discount rate, can be divided into three groups, namely coupons of level low, middle and high (denoted by 1,2,3 respectively).
%%% number of each coupon?
Another reason why we do not maintain the original types of coupons is that the number of records about offering each kind of coupons varies a lot.

\subsubsection{Preference Group Clustering}
%In our model, we assume that the users in one group have the same preference distribution.
To determine users with similar preference distribution (denoted by $P$), we cluster them into 4 different groups based on features only related to the merchant and users themselves.
%Specifically, we extract 19 features from \textsl{O2O Dataset} for clustering, such as the number of coupons each user receives from the merchant, the number of coupons each user uses, the distance between each user and the merchant, etc.
%%%number in each group?

\subsubsection{Strategy Group Clustering}
As we introduced before, our model considers the opponent adopts a stable strategy distribution, (denoted by $S$), for each user.
But in \textsl{O2O Dataset}, the number of records for each user is too small.
Therefore, we cluster users in each preference group into 10 different subgroups based on features only related to themselves.
These subgroups are called strategy group, and we assume the opponent adopts the same strategy distribution for users in each strategy group.
%Specifically, 9 extracted features are considered for clustering, such as ???.
%%% number of each group?

%In our PGM, in one group, we assume that one user $u$, may generate multiple consumption records, and the opponent $O$ adopt the stable strategy distribution to $u$.
%In this dataset, It is not usual to observe the multiple consumption records of one user, while users usually have one record.
%Therefore, we need to cluster the users in each group again, called strategy cluster, so that the $O$'s strategies to the users in each strategy cluster are close.
%For this goal, the features for clustering need be related to the merchant. Specifically, ??? what features.

\begin{algorithm}
        \caption{Preprocess of Coupon Usage Data}
        \label{alg:preprcess_real}
        \KwIn{Dataset $D$, company 1, number of preference group $N$, number of strategy group $M$}
        \KwOut{The preference distribution $\{P_1, \dots, P_N \}$, the strategy
         distribution $\{S_{1, 1}, \dots, S_{N, M} \}$ }
        $PG_{1}, \dots, PG_{N} \leftarrow$ Cluster the users in the $D$ according to their preference to company 1. \\
        \For{$i \leftarrow 1 \KwTo N$}{
            $SG_{i, 1}, \dots, SG_{i, M} \leftarrow $ Cluster the users in $PG_{i}$ according to the opponent's strategies.
            $P_i, S_{i, 1} , \dots, S_{i, M} \leftarrow LDA(\{ SG_{i, 1}, \dots, SG_{i, M} \})$
        }
\end{algorithm}

\subsection{Evalution}
In this subsection, we apply our model on the dataset and evaluate it by measuring its behavior prediction and inferred strategy distribution, to show that our model is effective.

\subsubsection{Behavior Prediction}
We first train our model on the training dataset, then we use our model to predict the behaviors of users in the testing dataset.
We evaluate our model on the measurement of negative log likelihood for prediction, compared with some baselines.

Mathematically, negative log likelihood is defined as
$$ \mathcal{L}(\theta) = -\sum \limits_{i=1}^N log(p(y_i | \theta, x_i))$$,
where $\theta$ is the model we want to evaluate, $N$ is the number of samples, $x_i$ is the features and $y_i$ is the ground truth of sample $i$, $p(y_i | \theta, x_i)$ is the output probability of $y_i$ from model $\theta$ when given $x_i$.
The smaller the likelihood of prediction is, the better the corresponding model is.
And in our experiments, $N = 7284$ for our model, as well as all the baselines.

%\paragraph{Baselines}
We consider 5 common the probabilistic prediction models as baselines:
\begin{itemize}
    \item Naive bayesian (NB) \cite{rish2001empirical}
    \item Logistic Regression (LR)
    \item Support Vector Machine (SVM) \cite{platt1999probabilistic}
    \item Random Forest (RF)  \cite{liaw2002classification}
    \item Neural Network (NN) \cite{lecun2015deep}
\end{itemize}
All the above baslines are implemented by sklearn \cite{pedregosa2011scikit}.
And their input features are the same extracted features we using for LDA.

\begin{table}[ht]
\caption{Comparison of Negative Log Likelihood}

\begin{center}
% \scriptsize
\begin{tabular}{|c|c|c|c|c|c|c|}
\hline
Model & NB & LR & SVM & RF & NN & LDA \\ \hline
Result & 494.93 & 580.26 & 1085.40 & 597.93 & 509.26 & \textbf{401.97} \\ \hline
\end{tabular}
\label{tab:likelihood}
\end{center}

\end{table}

As shown in Tab.\ref{tab:likelihood}, our model get the smallest negative log likelihood in prediction, meaning that it provides the best modeling for the real-world data.

\subsubsection{Distance of Strategy Distributions}
We also evaluate the distribution distance between our strategy distribution and the real strategy distribution.
The real strategy distribution for each strategy group is calculated by the number of coupons that all other merchants in the whole dataset provide to users in the group.
Similar to the preprocessing, these coupons are also divided into three groups as level low, middle and high according to their discount rate.
We adopt the Wasserstein Distance \cite{ramdas2017wasserstein} to measure the distance of two distribution, which is defined as
$$
W_1(\vec{p}, \vec{q}) = \operatorname{inf}\limits_{\pi \in \Gamma(\vec{p}, \vec{q})} \int_{\mathbb{R} \times \mathbb{R}} |x - y| \operatorname{d} \pi(x, y)
$$, where $\Gamma(\vec{p}, \vec{q})$ denotes the collection of all joint distributions on $\mathbb{R} \times \mathbb{R}$ whose marginals are $\vec{p}$ and $\vec{q}$ on the first and second factors respectively.

% , where $\vec{p} = (p_1, \dots, p_k)$ and $\vec{q} = (q_1, \dots, q_k)$
% are two distributions.
We consider two distributions as our baselines.
\begin{itemize}
    \item The overall distribution of received coupons. We count the total number of each kind of coupons that all other merchants in the whole dataset provide to all users of company 1 as the baseline distribution.
    \item Uniform distribution: $p = (\frac{1}{3}, \frac{1}{3}, \frac{1}{3})$, which is what a single merchant may assume for its opponent without knowing further information.
\end{itemize}

As shown in Tab.\ref{tab:distance}, the distance between our inferred distribution and the real distribution is the closest.

\begin{table}[ht]

\caption{Comparison of Distribution Distance}

\begin{center}
% \scriptsize
\begin{tabular}{|c|c|c|c|}
\hline
Model  & Uniform & Average  & LDA   \\ \hline
Result & 0.18794  & 0.13105   & \textbf{0.12303} \\ \hline
\end{tabular}
\label{tab:distance}
\end{center}
\end{table}

\subsection{Analysis}

Unlike the opponent's strategy that we can calculate from the dataset, the true preference distributions of users are hard to know.
Therefore we analyze the inferred distributions to show that they are reasonable to some extent.

\begin{figure}[ht]
    \center
    \includegraphics[width = 5in]{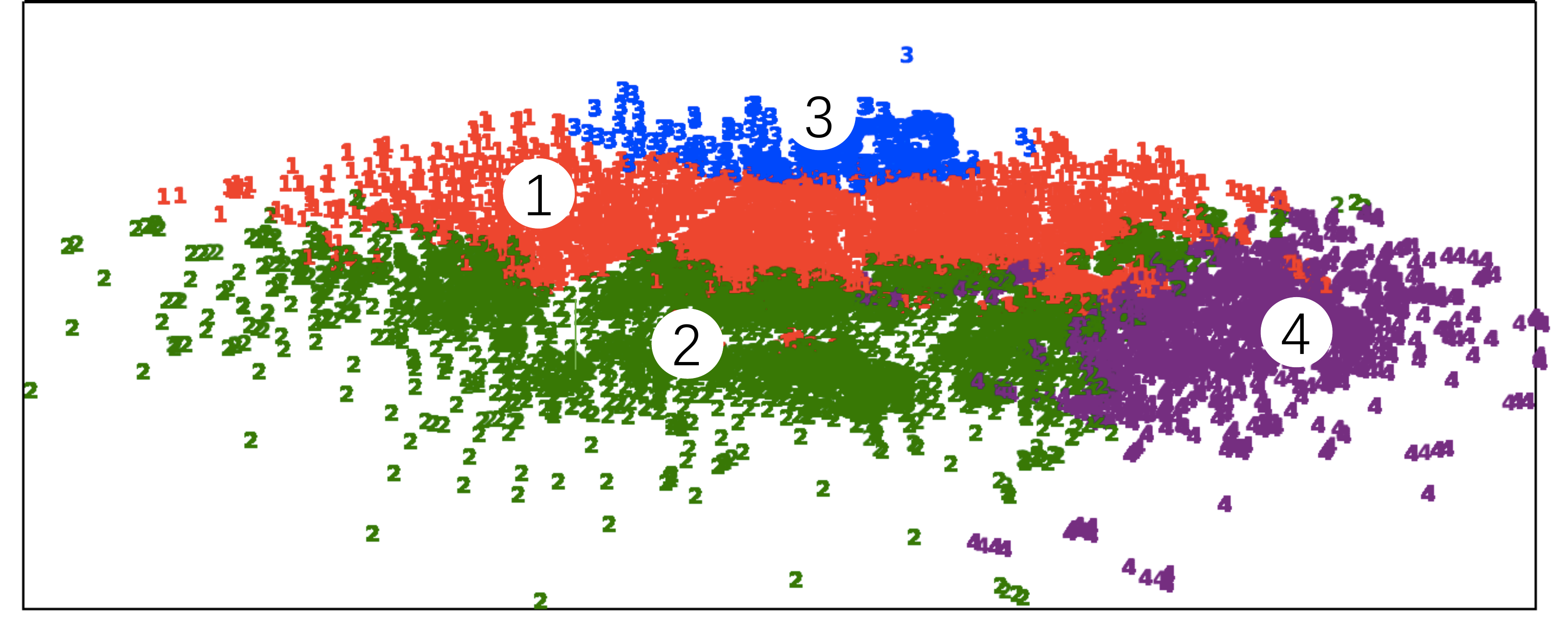}
    \caption{The visual results of our clustering for preference groups}
    \label{fig:tsne}
\end{figure}

Firstly, we show the visual results of our preference clustering in Fig. \ref{fig:tsne}.
The aim of preference clustering is to cluster the users with similar preference to one merchant. From Fig. \ref{fig:tsne}, we can see clearly that the users with the same preference are clustered together.

\begin{figure}
    \flushleft
    \includegraphics[width =5.5in]{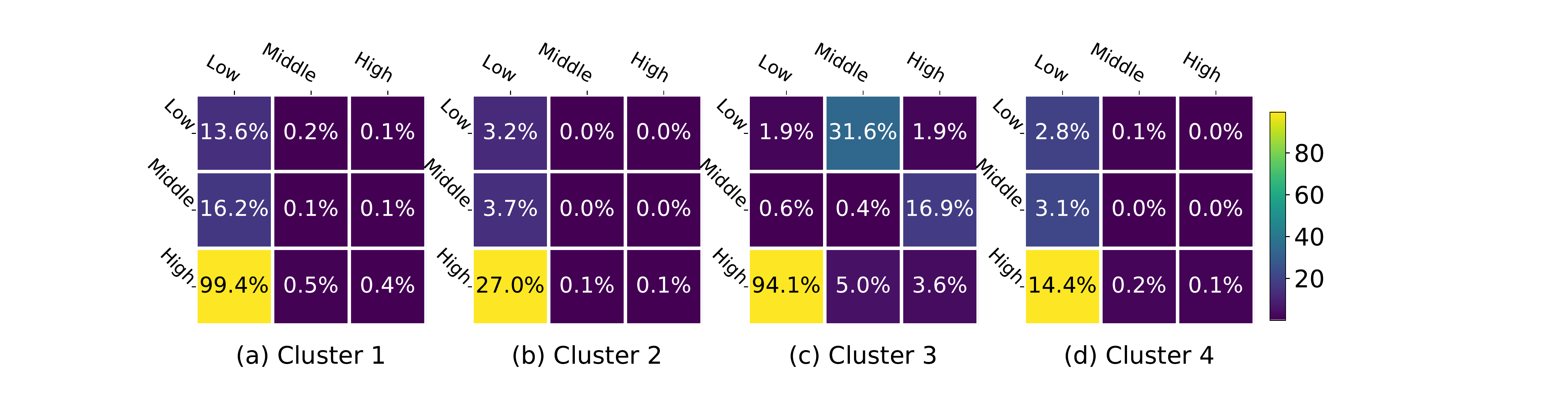}
    \caption{The heatmap of preference distributions of four clusters}
    \label{fig:pref_heatmap}

\end{figure}

%Through the Fig. \ref{fig:tsne}, the results of our clustering had satisfactory discrimination.
%Then with inference based on our model, we get the preference distributions and strategy distributions of each cluster.
Then, we show the preference distributions inferred by our model explicitly.
Fig. \ref{fig:pref_heatmap} shows the heat-map of preference distributions of four clusters.
The block ($i$, $j$) with lighter color represents when users receive ($i$, $j$) pair of coupons, the preference they choose company 1 is higher.
"Low", "Middle", "High" mean the effects of coupon respectively, as we introduced in Subsection. \ref{prepro}.
We can find that when company 1 chooses high coupons, the preference distribution of users is very high, close to 100\%.
These results confirm our intuition.

\begin{figure}
    \flushleft
    \includegraphics[width = 5.0in]{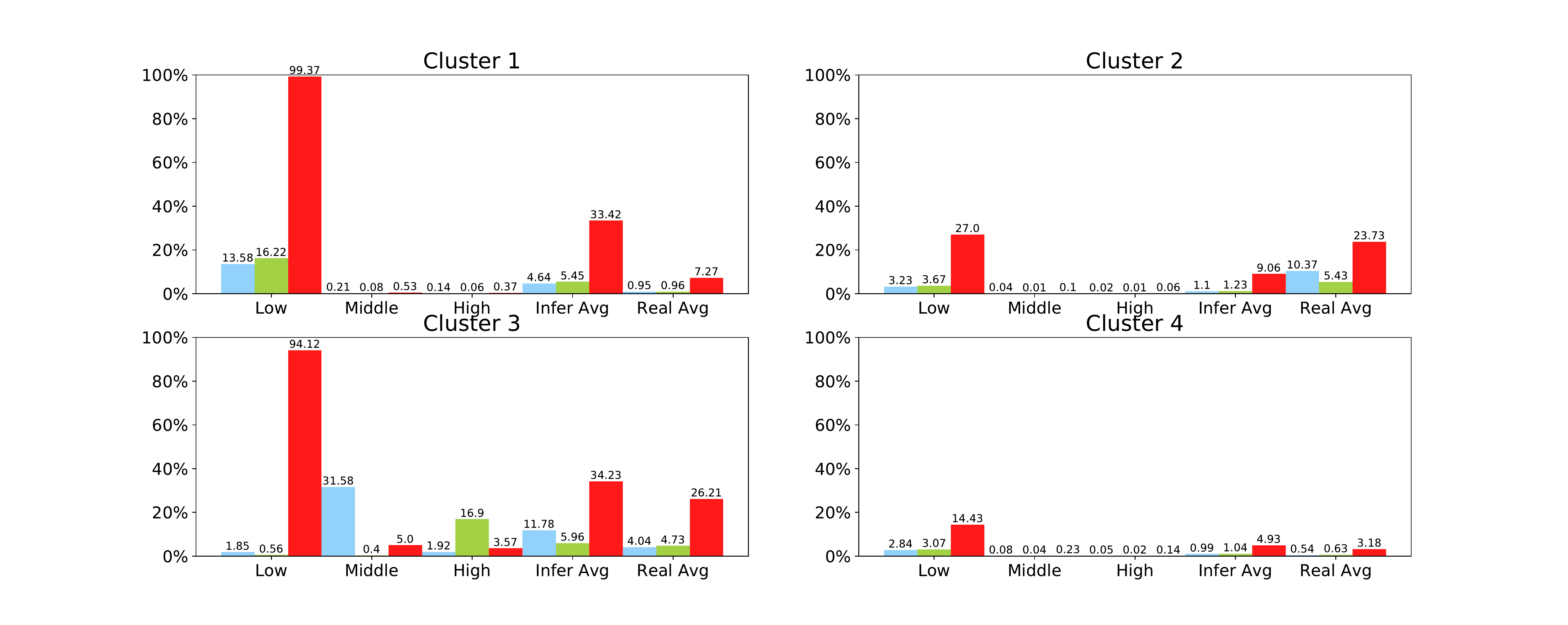}

    \caption{The histogram of preference distribution in each cluster.
    In each subfigure, the bars with the same colors mean that we adopt the same coupon.
    The 'Low', 'Middle', 'High' in the x-axis mean represent what coupon the opponent chooses respectively.
    The 'Infer Average' represent the mean preference regardless of the opponent's coupons when we choose the corresponding coupon.
    The 'Real Average' represent the average usage rates of the coupons regardless of the opponent's coupons when we choose the corresponding coupon.}

    \label{fig:pref_hist}
\end{figure}

We also plot the histogram of the distribution of preference in Fig.\ref{fig:pref_hist}.
The histogram of preference distribution in each cluster.
In each subfigure, the bars with the same colors mean that we adopt the same coupon.
The 'Low', 'Middle', 'High' in x-axis represent what coupons the opponent chooses respectively.
The 'Infer Average' represent the mean preference regardless of the opponent's coupons when we choose the corresponding coupon.
The 'Real Average' represent the average usage rates of the coupons regardless of the opponent's coupons when we choose the corresponding coupon.
It is easy to see that when we choose the high effective coupon and the opponent choose the low effective coupon, the preference to us is very high.
The most noticeable results in the Fig. \ref{fig:pref_hist} are the comparisons between 'Infer Avg' and 'Real Avg'.
We can find that if 'Real Avg' is higher than other clusters, the 'Infer Avg' is higher than other clusters too except the cluster 1, which is in line with the intuition.

% % The results of preference distributions on different coupons confirm
% Four figures represents four distributions in the four clusters.
% When we choose the good coupons (red bars), the preferences of the user to us are almost higher than other coupons.
% The average preference for the good coupon is always larger than the coupons. When the opponent chooses the good coupons, the preference for them is relatively higher than the others.
% The above analysis shows the output distribution from our model makes sense.

% In the end, we show some concrete transactions in the dataset.
% The records under the line are the real transactions in the dataset, while the opponents' coupons are results inferred by our model.
% We divide these transactions into the different group via the different pairs of coupons.
% It is easy to find that when we choose the high effective coupons and the opponent choose the low effective coupon, the users prefer to use our coupons.
% \begin{figure*}
%     \center
%     \includegraphics[width = 7.5in]{fig/infer_example.pdf}
%     \caption{Some examples of transactions from Coupon Usage Dataset. Each coloar indicate a different pair coupon.}
%     \label{fig:infer_example}
% \end{figure*}

\section{Practical Financial Scenario}\label{real}
In this section, we present our experiments on real financial data in two aspects.
\paragraph{Dataset}
We also evaluate our methods on the real business dataset from a company. The dataset is selected from its marketing records for its new service in September 2017, when coupons of different types were sent to customers to attract them to use the service. The dataset contains customers’ features (264 related attributes, such as one’s resident, age, gender and so on), types of their received coupons, and their response (whether they used the service) in the next 15 days. The ratios of the positive and negative samples is close 1 : 200.

\paragraph{Likelihood On Real Data}
Similar to Section \ref{o2o_data}, we apply our model to this dataset, infer hidden variables from training data and make predictions on the testing data accordingly.
We calculated its negative log-likelihood and compared it with those got from other baselines.

% \textbf{Dataset} We used the consumption data of a service provided by an Internet company. The record for each user consists of three parts: the type of coupon he receives, basic features of him, and the amount he used the service within 15 days after receiving the coupon.
% The types of coupons are distinguished by their ids and features are normalized into several integer intervals.
% %This data set will be open to the public in the future.
\begin{table}[ht]

\caption{Comparison of Likelihood}

\begin{center}\scriptsize
\begin{tabular}{|c|c|c|c|c|}
\hline
Model & NN & NB & LR & LDA \\ \hline
Result & 449.52 & 701.56 & 1432.85 & 259.60\\ \hline
\end{tabular}

\label{likelihood}
\end{center}
\end{table}

As shown in Table~\ref{likelihood}, 10000 records from the dataset are used, among which 90\% is used for training and the rest for testing.
% For our method, the data of basic features is used for clustering users and opponent's strategy as well as users' preference is inferred for each group of clustered users.
% For DNN, Naive Bayes and Logistic Regression, the basic features and type of coupon are used as features while the borrowed money be the labels.
The negative log likelihood of our method's predictions on testing data is significantly smaller than other models, which means our method captures the users' behaviors much better.

\paragraph{Practical Financial Marketing}
With the help of the company, we tested our method on practical marketing for the same service.
Similar to the process as the simulation experiments, we applied the DRL framework for Price War with the information learning from the Price War LDA (Fig. \ref{fig:framework}) to practical financial marketing.
In December 2017, 600,000 customers of the company were divided randomly into three groups of the same size.
Our framework decides awards for customers from one group, while Random Strategy and the standard DQN for other two respectively as baselines.
Within two weeks after receiving awards, 2.8053\% customers from the first group use the product, while 2.1697\% of the second and 2.4966\% of the third.

Compared with the usage rate of Random Strategy, our method got an improvement of 29.29\% on usage rate, while DQN only got 15.07\%. This means our method, considering opponent's strategy and customer's preference, do improve the effect of personalized marketing compared with those do not considering them.

\section{Conclusions}

In this paper, we formalize the Internet price war as an imperfect and incomplete information game.
We design an LDA to explore unknown variables from one participant's perspective.
The inferred information is shown to help decision making method, like DRL and DP, for finding better strategies in simulated experiments.
And the model also exhibits better characterization for an open dataset from a practical business.
% Our method performs great enhancement to DRL, proved through a series experiment.
It is the first time that LDA is used in a game scenario and makes efforts in the competitive business environment.
This design not only makes a major contribution towards achieving better market sharing in an Internet price war but also inspire a novel technique for dealing with incomplete and imperfect information games.

%\section*{Acknowledgments}

%This research was partially supported by the National Nature Science Foundation of China (No. 11426026, 61632017, 61173011), by a Project 985 grant of Shanghai Jiao Tong University and by Ant Financial.

%\appendix

%\newpage
% \clearpage
\bibliographystyle{plain}
\bibliography{reference}

\end{document}